\documentclass[10pt,a4paper]{article}

\usepackage{newtxtext}   

\usepackage{titlesec}
\titleformat*{\section}{\bf\normalsize}
\titleformat*{\subsection}{\normalsize}

\titlespacing{\section}{0pt}{2ex}{1ex}
\titlespacing{\subsection}{0pt}{2ex}{1ex}
\titlespacing{\subsubsection}{0pt}{2ex}{1ex}

\usepackage{fancyhdr}
\usepackage[utf8]{inputenc}
\usepackage[top=2cm,bottom=2cm,left=2cm,right=2cm]{geometry}
\usepackage[usenames,dvipsnames]{color}

\usepackage{hyperref}
\hypersetup{bookmarks=true,pdfborder={0 0 0},colorlinks=true,urlcolor=blue,linkcolor=blue,citecolor=blue}
\hypersetup{pdftitle={A Neural Emulator for Uncertainty Estimation of Fire Propagation}}
\hypersetup{pdfauthor={Andrew Bolt, Conrad Sanderson, Joel Janek Dabrowski, Carolyn Huston, Petra Kuhnert}}

\usepackage{amsmath}
\usepackage{amssymb}
\usepackage{amsfonts}
\usepackage{algorithmic}
\usepackage{graphicx}
\usepackage{booktabs}
\usepackage{textcomp}
\usepackage{xcolor}
\usepackage{makecell}
\usepackage{gensymb}   

\usepackage{enumitem} 
\usepackage{multirow}

\usepackage[british]{babel}
\usepackage[none]{hyphenat}
\usepackage{caption}
\newcommand*\mean[1]{\overline{#1}}

\begin{document}

\renewcommand{\baselinestretch}{1.05}\small\normalsize

\thispagestyle{empty}
\pagestyle{empty}

\begin{minipage}{1\textwidth}
\centering

\noindent
\mbox{\scalebox{1.6}{\textbf{A Neural Emulator for Uncertainty Estimation of Fire Propagation}}}

\vspace{5ex}

\large
\noindent
Andrew~Bolt\textsuperscript{{\tiny~}$\dagger$},
Conrad~Sanderson\textsuperscript{{\tiny~}$\dagger\ddagger$},
Joel~Janek~Dabrowski\textsuperscript{{\tiny~}$\dagger$},
Carolyn~Huston\textsuperscript{{\tiny~}$\dagger$},
Petra~Kuhnert\textsuperscript{{\tiny~}$\dagger\diamond$}\\
~\\
\normalsize
\textsuperscript{$\dagger$}{\tiny~}\textit{Data61 / CSIRO, Australia;}~
\textsuperscript{$\diamond$}{\tiny~}\textit{Australian National University, Australia;}~
\textsuperscript{$\ddagger$}{\tiny~}\textit{Griffith University, Australia}
   
\end{minipage}

\parskip=0ex
\parindent=3ex

\vspace{2ex}
\section*{Abstract}

Wildfire propagation is a highly stochastic process where small changes in environmental conditions
(such as wind speed and direction)
can lead to large changes in observed behaviour.
A traditional approach to quantify uncertainty in fire-front progression
is to generate probability maps via ensembles of simulations.
However, use of ensembles is typically computationally expensive,
which can limit the scope of uncertainty analysis.
To address this, we explore the use of a spatio-temporal neural-based modelling approach 
to directly estimate the likelihood of fire propagation given uncertainty in input parameters. 
The uncertainty is represented by deliberately perturbing the input weather forecast during model training.
The computational load is concentrated in the model training process,
which allows larger probability spaces to be explored during deployment.
Empirical evaluations indicate that the proposed model achieves comparable fire boundaries
to those produced by the traditional SPARK simulation platform,
with an overall Jaccard index (similarity score) of 67.4\% on a set of 35 simulated fires.
When compared to a related neural model (emulator)
which was employed to generate probability maps via ensembles of emulated fires,
the proposed approach produces competitive Jaccard similarity scores while being 
approximately an order of magnitude faster.

\vspace{3ex}
\hrule
\vspace{1ex}

\section{Introduction}

Wildfires are a destructive phenomenon affecting natural flora and fauna
as well as posing many risks to human life and property.
Response and management efforts are critical to mitigating the damage caused by wildfires~\cite{bot2022systematic,Bowman2020VegetationFI,scott2013fire}.
To this end, in silico wildfire simulations are often performed
in order to aid managers in making informed choices
for fire risk management and mitigation efforts, as well as evacuation planning.

There are a number of computational fire spread models that are currently deployed for 
simulating and predicting the spreading behaviour of fires~\cite{finney1998farsite,Spark2015,Phoenix2008}.
These models typically take into account underlying empirical observations of fire spread given specific conditions,
such as
fuel class (eg.~vegetation type),
terrain characteristics (eg.~slope),
as well as weather (eg.~temperature, wind speed and direction).
These simulations are usually deterministic~\cite{Spark2015,Phoenix2008},
and predictions depend on static initial conditions (initial fire location, fuel class, terrain maps)
and estimates of future weather conditions.

As wildfire spread is inherently a stochastic process,
a single simulation of fire progression is unlikely to account for \textit{uncertainty}
in the underlying variables driving the model. 
For example, there are variabilities within fuel classes,
uncertainty in classifying a fuel,
and uncertainty in weather forecasting.
As such, it is desirable to produce likelihood estimates of fire location (probability map) at a given point in time,
in order to provide a form of uncertainty quantification.

One approach for dealing with uncertainty in a given variable is to run an ensemble of simulations,
where each simulation draws the value of the variable from a distribution.
Ensemble predictions are a standard approach in areas such as weather and flood forecasting~\cite{gneiting2005, leutbecher2007, wu2020}.
However, a major limitation of ensemble-based simulations is that 
the parameter space increases exponentially with the number of variables.
The size of the ensembles is hence constrained by the computational demands of the simulations,
which in turn can limit the scope of uncertainty analysis to a subset of variables.

To address this shortcoming, in this paper
we explore the use of a spatio-temporal neural-based model
to \textit{directly} estimate the output of an ensemble of simulations.
The proposed model is trained via probability maps,
which are generated from ensembles of fire-fronts produced by the traditional SPARK simulation platform~\cite{Spark2015}.
Within the training ensembles, weather parameters are perturbed according to a noise model. 
The computational load is hence concentrated in the model training process, rather than during deployment.
Empirical evaluations show that the proposed model
generalises well to various locations and weather conditions,
and generates probability maps comparable to those \textit{indirectly} generated
via ensembles of SPARK simulations.

The paper is continued as follows.
Section~\ref{sec:related_work} provides a brief overview of related work.
Section~\ref{sec:modelling} describes the proposed model architecture.
Section~\ref{sec:evaluation} provides comparative evaluations.
We conclude by discussing the results, limitations and future work in Section~\ref{sec:conclusion}.

\newpage
\section{Related Work}
\label{sec:related_work}

Recent work has investigated use of machine learning methods for wildfire related applications,
covering topics such as
susceptibility prediction,
fire spread prediction,
fire ignition detection and decision support models~\cite{abid2021survey,bot2022systematic,jain2020review}.
Deep learning architectures have also been used to develop models related to wildfires~\cite{allaire2021,burge2020convolutional,hodges2019wildland,radke2019firecast}.
Burge et al.~\cite{burge2020convolutional} and Hodges et al.~\cite{hodges2019wildland}
present convolutional neural network (CNN) emulators that estimate fire spread behaviour,
while Allaire et al.~\cite{allaire2021} present a CNN emulator for hazard assessment within a target region.

Neural networks have also been used to predict the likelihood of a region being burned,
given the confidence the model has in its own prediction.
Radke et al.~\cite{radke2019firecast} propose a CNN based emulator
that predicts the likelihood a pixel is burned within a 24 hour period.
Sung et al.~\cite{sung2021} also develop a model that predicts the likelihood that a pixel is burned;
the model is trained using daily fire perimeters, rather than synthetic data.
Dabrowski et al.~\cite{Janek_2023} combine a Bayesian physics informed neural network
with level-set methods~\cite{mallet2009modeling} for modelling fire-fronts.
Outside of the fire space, Gunawardena et al.~\cite{gunawardena2021}
deploy a hybrid linear/logistic regression based surrogate model
to estimate the behaviour of ensemble weather modelling.
Egele et al.~\cite{Egele_ICPR2021} propose 
an automated approach for generating an ensemble of neural networks to aid uncertainty quantification.

In a separate line of research,
surrogate models (also known as model emulators)
can be used as more computationally efficient representations
of complex physical process models~\cite{alizadeh2020managing,sanderson2021emulation}.
Such models are often based on statistical or machine learning approaches.
Models of various physical systems have been achieved using techniques such as logistic regression~\cite{gunawardena2021},
Gaussian processes~\cite{bastos2012,conti2009},
random forests~\cite{gladish2018,leeds2013}
and deep neural networks~\cite{Kashinath2021,kasim2021,sanderson2021emulation,sit2020}. 

Potential advantages of neural network based model emulators include
accelerated computation through the use of Graphics Processing Units (GPUs)
via open-source packages such as TensorFlow~\cite{tensorflow_usenix_2016} and PyTorch~\cite{PyTorch_NeurIPS_2019}.
Furthermore, since neural network models are in essence a large set of numerical weight parameters,
they are portable across computing architectures,
easily deployed in cloud-based computing environments,
and can be integrated into larger frameworks with relative ease~\cite{sanderson2021emulation}.

\section{Model Architecture}
\label{sec:modelling}

\begin{figure*}[!b]
\vspace{-2ex}
\begin{minipage}{1\textwidth}
\centering
\hrule
\vspace{1ex}
\begin{minipage}{0.65\textwidth}
  \centering
  \includegraphics[width=1.0\textwidth]{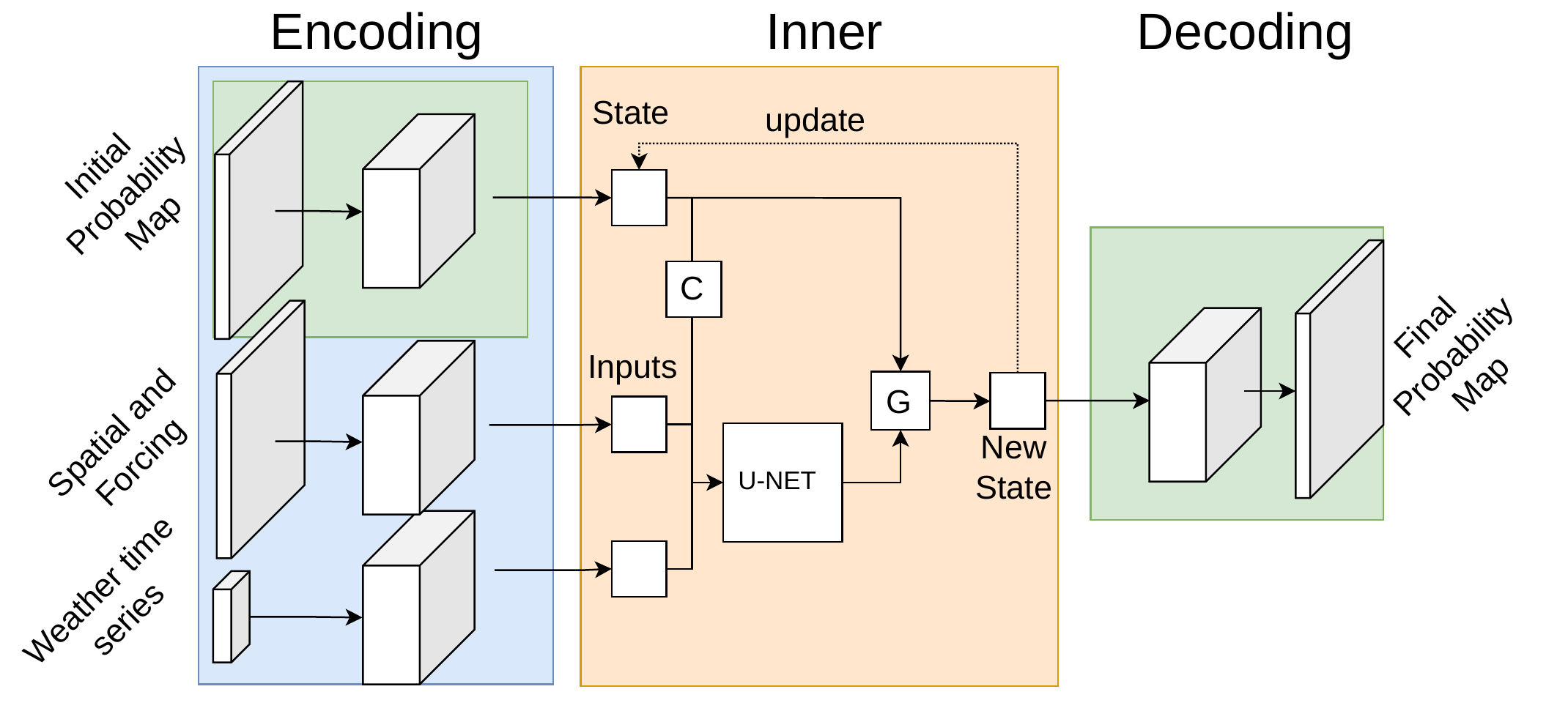}
\end{minipage}
\hfill
\begin{minipage}{0.34\textwidth}
    \caption
    {
    \small
    The proposed model architecture based on~\cite{bolt2022}.
    Key additions are in the inner component:
    edge detection (layer C)
    and 
    constrained update layer (layer G).
    The initial probability map and features are encoded via down-sampling.
    The inner component of the model updates the latent probability map using spatial features and weather variables
    (see Fig.~\ref{fig:arch_inner} for details).
    Once all weather inputs have been used a final latent probability map estimate is produced and decoded via up-sampling.
    The layers used to encode and decode probability maps are trained separately as an autoencoder
    (see Fig.~\ref{fig:autoencoder} for details).
    }
    \label{fig:arch_overview}
\end{minipage}
\end{minipage}

\hrule
\vspace{2ex}

\begin{minipage}{1\textwidth}
\begin{minipage}{0.65\textwidth}
  \centering
  \includegraphics[width=0.85\columnwidth]{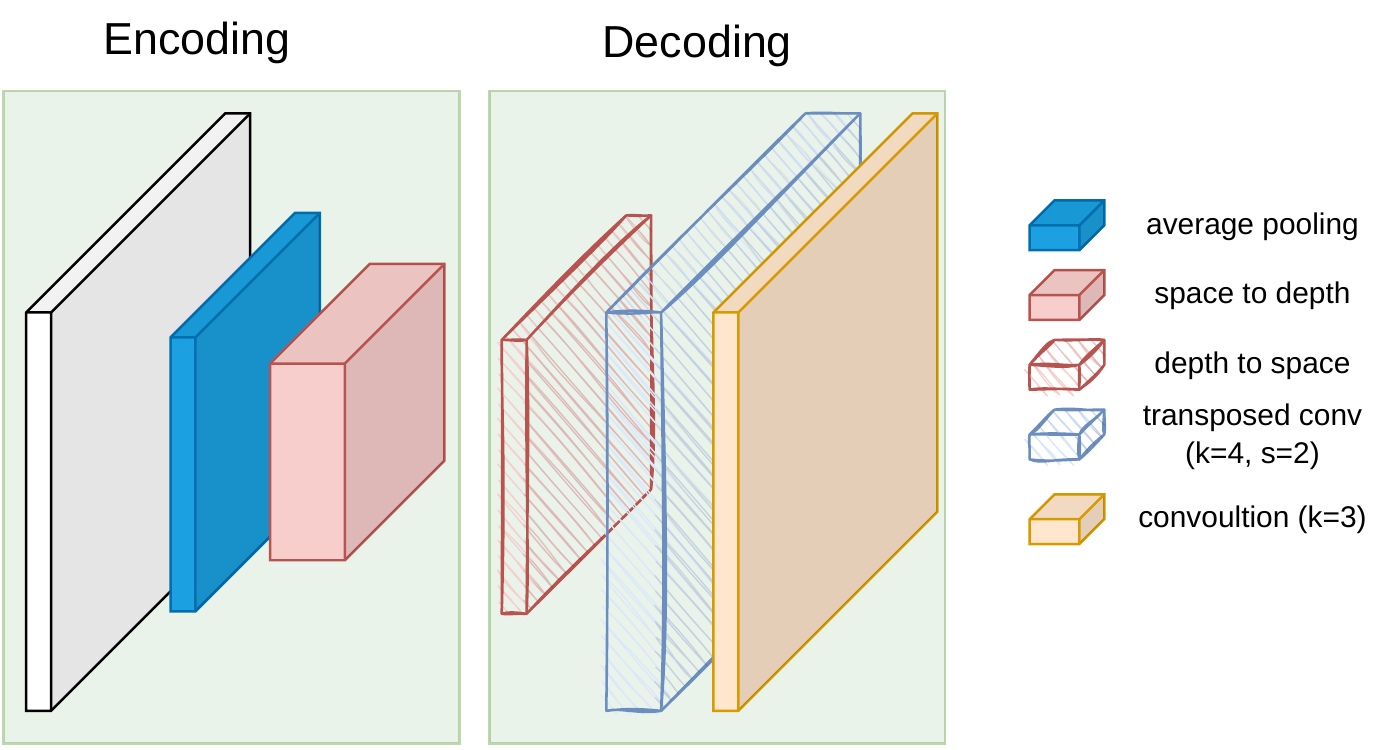}
\end{minipage}
\hfill
\begin{minipage}{0.34\textwidth}
  \caption
    {
    \small
    The autoencoder component of the model.
    This component is trained separately on probability maps,
    and the weights are frozen when incorporated into the main model.
    The downscaling component uses only pooling and space to depth operations
    to preserve probability values within the latent representation.
    }
    \label{fig:autoencoder}
\end{minipage}
\end{minipage}

\vspace{2ex}
\hrule
\vspace{2ex}

\begin{minipage}{1\textwidth}
\begin{minipage}{0.7\textwidth}
  \centering
  \includegraphics[width=1\textwidth]{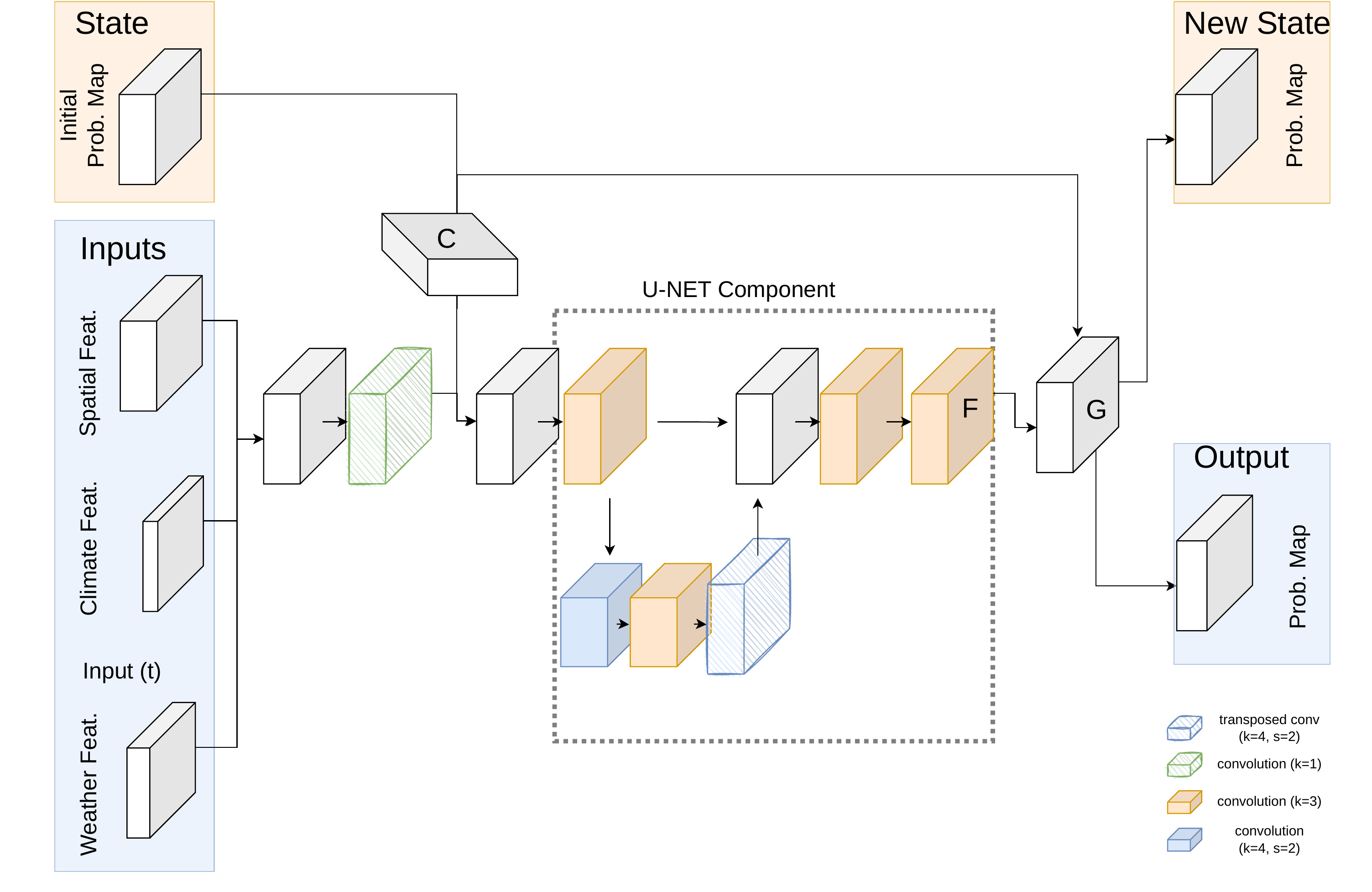}
\end{minipage}
\hfill
\begin{minipage}{0.29\textwidth}
  \caption
    {
    \small
    The inner component of the model.
    The latent spatial, climate, and weather time-series are concatenated.
    The latent arrival state is concatenated with this stack
    and is passed through a shallow U-Net~\cite{ronneberger2015}.
    The resulting output is used as a residual component.
    This residual is added to the latent arrival state, which is then used in the next time step.
    When the weather inputs are exhausted, the final latent arrival state is the output.
    }
  \label{fig:arch_inner}
\end{minipage}
\end{minipage}
\end{figure*}

We extend the spatio-temporal neural network model architecture presented in~\cite{bolt2022}.
As illustrated in Fig.~\ref{fig:arch_overview},
the architecture comprises an \textit{Encoding} block,
an \textit{Inner} block, and a \textit{Decoding} block.
Compared to~\cite{bolt2022},
the inner block is expanded with two layers:
layer $C$ implements edge detection,
and
layer $G$ incorporates update constraints.

There are three types of inputs to the model:
(i)~initial probability map,
(ii)~spatial data and forcing terms,
(iii)~weather time series.
The probability map consists of an image where pixels values represent the likelihood
of the fire reaching a location at a given point in time.
The spatial data are images representing height maps and land classes (eg.~forest, grassland).
The forcing terms are curing and drought factors.
The weather time series consists of the mean and standard deviations of temperature,
wind ($x$ and $y$ components), and relative humidity.
These are summary statistics of the values used in the ensemble of simulations used for training.

The encoder and decoding steps are as described in~\cite{bolt2022}.
The encoding step is used to create a more compact latent representation of features.
This is illustrated by the light blue shading in Fig.~\ref{fig:arch_overview}.
The probability map is encoded using an autoencoder, illustrated in Fig.~\ref{fig:autoencoder}.
The relationship between pixel value and probability is preserved in the encoding process.
The autoencoder component of the model is trained separately
and its weights are frozen during the main training step.
This transfer learning approach ensures that dynamics are not inadvertently learned by the encoding/decoding layers,
as well as reducing the number of weights that need to be trained by the full model.
The height map and landclass map features are encoded
using alternating convolutional and strided convolutional layers which reduce the spatial extent.

The inner model is designed to handle the dynamics of the system, 
incorporating a shallow U-Net structure~\cite{ronneberger2015,wang2019}.
See Fig.~\ref{fig:arch_inner} for details.
New latent probability map estimates are generated,
which are then propagated forward in time as new weather values are processed.
The updates are also informed by the latent spatial and climate features.

\begin{figure}[!t]
  \begin{minipage}{0.6\textwidth}
    \centering
    \begin{minipage}{0.40\columnwidth}
      \centering
      \includegraphics[width=0.9\textwidth]{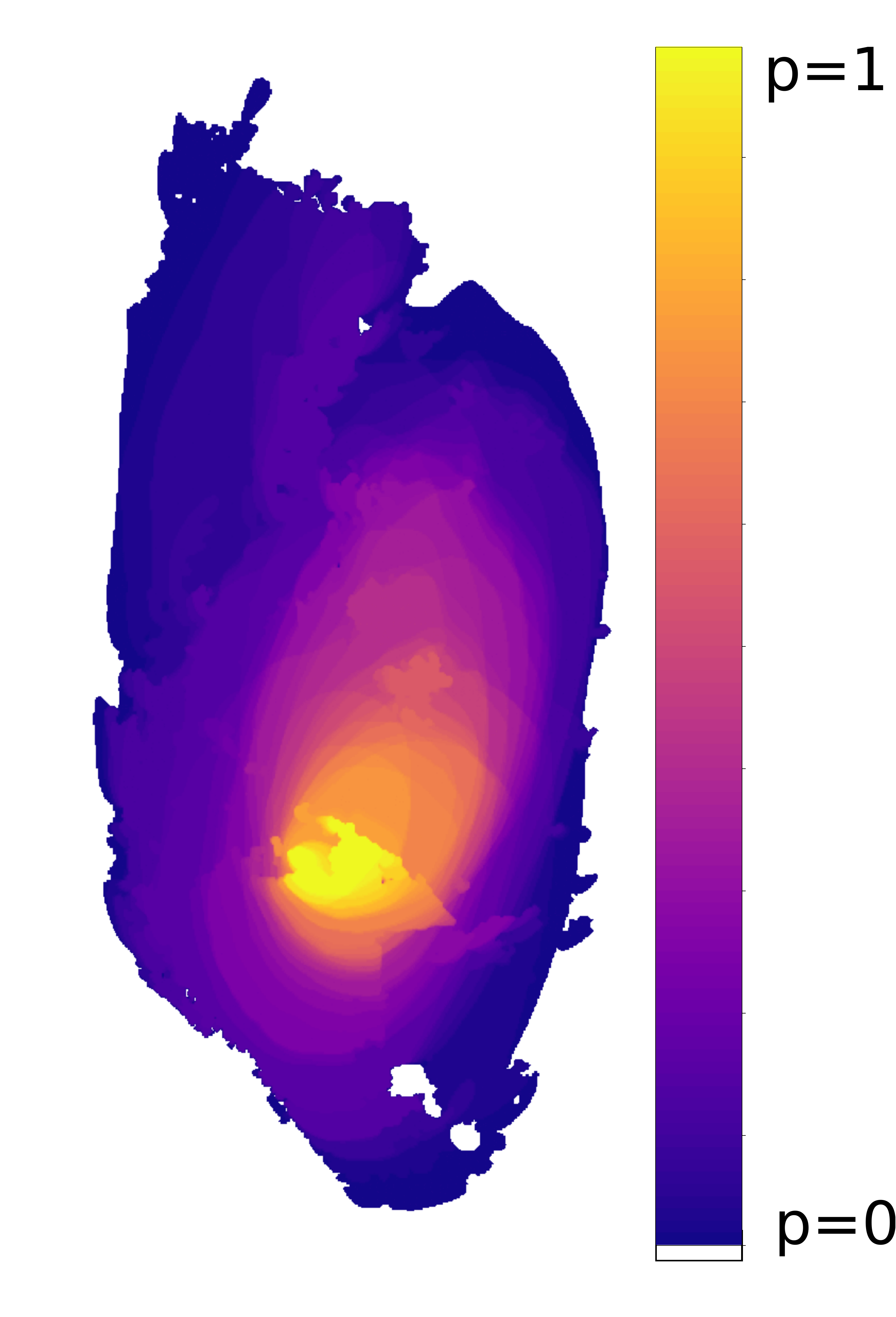}\\
      \footnotesize\textbf{(a)}
    \end{minipage}
    \hfill
    \begin{minipage}{0.40\columnwidth}
      \centering
      \includegraphics[width=0.9\textwidth]{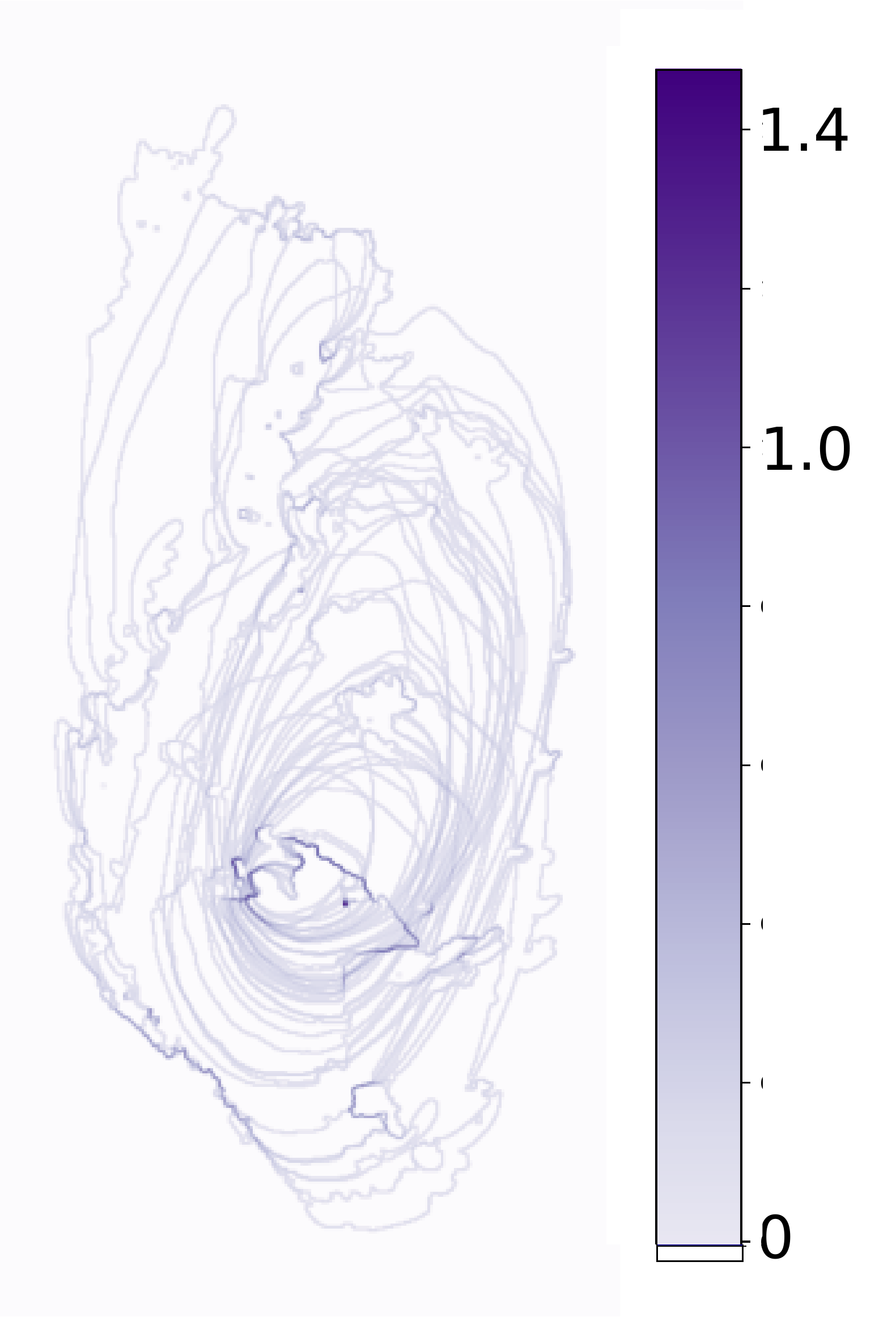}\\
      \footnotesize\textbf{(b)}
    \end{minipage}
    \vspace{-0.5ex}
    \caption
      {
      \small
      \textbf{(a)}:
      Probability map (non-latent) of fire spread generated from an ensemble of 50 fire simulations.
      14 hours have elapsed from the ignition time.
      The likelihood of a pixel being in the burned state
      is determined by the fraction of simulations in which a pixel is burned.
      Yellow indicates high likelihood, while blue indicates low likelihood.
      White indicates no burning in any simulation.
      Individual fire fronts within the ensemble lie on contours of the probability map.
      \textbf{(b)}:
      The magnitude of Sobel edges for the corresponding probability map.
      Gradients between probability contours correspond to individual fire fronts within the ensemble;
      these can be used as sources of fire growth.
      }
    \label{fig:ensemble_example_and_edges}
    \vspace{1ex}
  \end{minipage}
  \hfill
  \vline
  \hfill
  \begin{minipage}{0.35\textwidth}
    \centering
    \vspace{2.4ex}
    \includegraphics[width=0.9\textwidth]{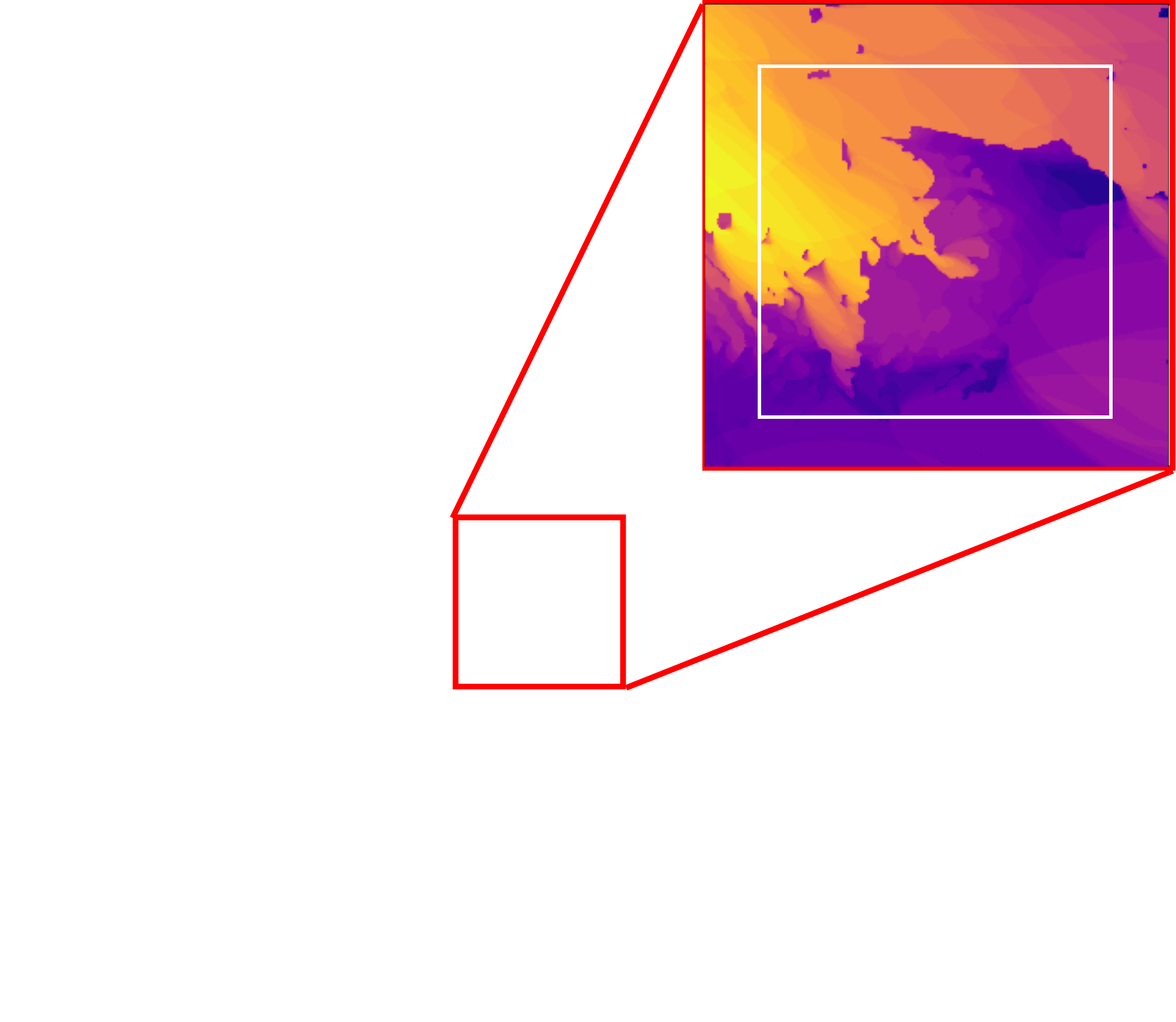}\\
    \caption
      {
      \small
      An example of a cropped image suitable for training.
      The original image is 1792$\times$2048.
      The cropped region is 256$\times$256 shown by the red box. The area outside the white box indicates the 32 pixels used for padding.
      }
    \label{fig:cropping}
    \vspace{4.5ex}
  \end{minipage}
  \hrule
  \vspace{-1ex}
\end{figure}

Within the inner model, layer $C$ applies a Sobel edge filter~\cite{Danielsson_1990},
where the output is $x$ and $y$ edge components.
The edge detection aims to make it easier for the neural network
to identify where fire fronts are within the probability map.
This approach has parallels with saliency guided training~\cite{Ismail_NEURIPS2021}.
Fig.~\ref{fig:ensemble_example_and_edges} shows examples of edge magnitudes.

Layer $G$ implements constrained updates to the latent probability map.
A new map $P^{~t+1}$ at time $t+1$ is generated using the previous map $P^{~t}$
and update terms from $F^{~t}$ at time $t$,
where $F$ is the output layer from the U-Net component of the model (see Fig.~\ref{fig:arch_inner}).
$F$ uses a sigmoid activation function so that values lie on the $[0, 1]$ interval~\cite{chollet2015keras}.

More specifically, layer $G$ acts to update the latent probability map as follows.
Let $F_{ij}^{~t}$ denote the value of the $i$-th kernel of $F$ at pixel location $j$ at time $t$.
Let $P^{~t}_j\in [0,1]$ be the probability that pixel $j$ has been burned by time $t$.
The model is trained such that $F_{ij}^{~t}$ terms estimate the likelihood that fire spreads
from a surrounding local region to location $j$ at time $t+1$.
The spread characteristics and likelihoods are learned by the model in the U-Net component.
The probability that pixel $j$ has been burned at time $t+1$ is estimated using:

\noindent
\begin{equation}
  P^{~t+1}_{j} = G(P^{~t}_j, F^{~t}_j) = 1 - \left( 1 - P^{~t}_j \right) \prod\nolimits_i \left( 1 - F^{~t}_{ij} \right)
  \label{eq:prob_update}
\end{equation}

\noindent
In the above equation, the updated likelihood is given by 1 minus the probability
that the pixel is not burned by an exterior source $\left( 1 - F^{~t}_{ij} \right)$,
and is not already burned $\left( 1 - P^{~t}_j \right)$.
The update satisfies the following three constraints:
\begin{enumerate}[noitemsep,topsep=4pt]
\item
The likelihood a pixel is burned is non-decreasing and bounded by 1,
ie., ~$1 \geq P^{~t+1}_j \geq P^{~t}_j$.
 
\item
The likelihood a pixel is burned is unchanged if there are no external sources,
ie., ~$P^{~t+1}_j \rightarrow P^{~t}_j$ \text{if} $\forall~i: F^{~t}_{ij} \rightarrow 0$.

\item
If there is at least one source that is very likely to burn a pixel,
then the likelihood the pixel is burned approaches~1,
ie., ~$P^{~t+1}_j \rightarrow 1$ if $\exists~i~\text{s.t.}~F^{~t}_{ij} \rightarrow 1$.
\end{enumerate}

We have empirically observed that in order for the model to converge well in training,
it is necessary to ensure that the $F_i$ terms are sufficiently small
so that the latent probability map is gradually updated between iterations.
More specifically, preliminary experiments indicated that the model is unlikely to converge during training
when using the default value of $0$ for the initial bias of layer $F$.
As such, we have empirically set the initial bias to~$-5.0$.

\clearpage

\section{Empirical Evaluation}
\label{sec:evaluation}

\subsection{Data Preparation}
\label{subsec:dataset}

We use a dataset comprised of 180 fire probability maps,
split into 145 training and 35 validation samples.
Likelihood estimates are given in half hour intervals throughout the duration of the sample.
To generate each probability map, we produce an ensemble of 50 fire perimeter simulations using the SPARK platform~\cite{Spark2015}.
For each time interval, the fire perimeters are calculated and the ensemble is flattened into a probability map.
See Fig.~\ref{fig:ensemble_example_and_edges} for an example.

The location of the fires is representative of regional South Australia.
Realistic
land classification maps%
\footnote{Land classification datasets derived from Department of Agriculture and Water Resources
\textit{Land Use of Australia 2010-11} dataset.}
and
topology data%
\footnote{Topography data sets derived from Geoscience Australia SRTM-derived
1~Second Digital Elevation Models Version 1.0.}
are used.
There is a large bias in land class distribution towards
temperate grassland (55.4\%),
sparse shrubland (11.3\%),
water/unburnable (10.6\%)
and mallee heath (10.1\%).
Other classes represent 12.6\% of the total area.
Weather conditions\footnote{Meteorological time series data sets derived from Australian Bureau of Meteorology automated weather station data.}
are sampled from days where the \textit{forest fire danger index} is severe~\cite{dowdy1975,mcarthur1967};
the temperatures are typically high and the relative humidity is low.

The fires within an ensemble are identical,
except that each time series of weather values is perturbed using the following noise model. For proof-of-concept we use a straightforward random walk noise model.
At each time step (30 min interval), $\pm 3\%$ of the mean variable value is added.
We choose $3\%$ based on preliminary analysis;
this value is large enough so that simulations within an ensemble display distinctly different features,
while being small enough that the general shape of each fire is similar.
Let $x_i$ be the noise added at time $i$,
$V_j$ the noiseless variable from the time-series $V$ at time $j$.
We then define {$\widetilde{V}_j = V_j + \sum\nolimits_{i=1}^j x_i$} as a noisy weather source,
where {\small $x_i = \text{RAND}(\{-1, 1\}) \times 0.03 \times \mean{V}$},
and $\mean{V}$ is the mean value of $V$.
Each simulation within an ensemble uses an independently calculated $\widetilde{V}$.

The weather variables form the time series and are wind speed, wind direction, temperature and relative humidity.
Due to use of degrees for wind direction within the weather data,
the wind direction was perturbed by $\pm 2\degree$ at each step,
rather than $3\%$ of the mean value%
\footnote{Percentage of the mean value cannot be used here,
due to the nature of the measurement used for wind direction (degrees).
If the mean value was used, winds with mean direction of $0\degree$ would never be perturbed,
while winds with mean direction of $359\degree$ would have a large perturbation.}.

There are $N=50$ unique time series for each variable (eg. temperature) of length $T$.  
Rather than designing the model to use $N$ inputs at each time step for each variable,
we use their mean and standard deviations, taken across the ensemble at a given time.
This produces an input time series of size $(2, |V|, T)$, where $|V| = 4$ is the number of weather variables

Several pre-processing steps are applied to the features before they are presented to the model.
The sample bounds, coordinate reference system, and resolutions (30m) of all spatial data are all made identical.
Height maps are converted into gradient maps.
The wind components of the weather data are converted from wind speed and direction to $x$ and $y$ components.
Normalisation of weather and forcing terms is performed so that values fall within the range $-1$ to $+1$.

\subsection{Training}
\label{subsec:metrics_and_loss}

The model training process performs further modification to the data, as follows.
When a sample is selected,
a~random time within the burn duration is chosen as the initial probability map.
The final probability map is taken $D$~intervals later,
and the corresponding weather variables are extracted. 

Once a starting time is selected, a random square of the probability map is cropped,
such that the central pixel $j$ has a value, $x_j \in (0, 1)$, ie., excluding 0 and 1.
We center the cropping on this pixel
so that the cropped region is likely to contain pixels that have an intermediate probability of being burned.
Such regions are likely to undergo dynamic changes within the time interval $D$ used for training.
We have empirically observed that in regions where pixel values are dominated by 0 or 1,
the values are less likely to change during the time interval $D$.
Fig.~\ref{fig:cropping} shows an example of a cropped initial probability map suitable for training.

Each sample then undergoes a random rotation and/or reflection operation.
This, as well as randomly choosing a starting time and a cropping location,
is used for data augmentation with the aim of preventing over-fitting.
The reduction in size reduces memory requirements,
while the standardisation of sizes allows training to be batched. 

The autoencoder component is trained using mean squared error (MSE) loss. 
For training the model, a loss function is used that evaluates how well the model reduces the MSE
between predicted ($y_p$) and target ($y_t$) fire shapes
compared to the MSE loss between target and initial ($y_0$) fire shapes.
Let MSE$(a, b)$ be the pixelwise mean squared error between images $a$ and $b$.
The loss $\mathcal{L}$ of fire state~$\mathcal{P}$ is defined as
$\mathcal{L} (\mathcal{P})
=
\mathrm{log}_{10}
\left(
\Big\{ \mathrm{MSE}(y_p, y_t) + \tau \Big\}
/
\Big\{ \mathrm{MSE}(y_0, y_t) + \tau \Big\}
\right)$.
The term $\tau = 10^{-12}$ is used to remove singularities that arise if either MSE value approaches zero.

Models were implemented using TensorFlow~\cite{tensorflow_usenix_2016},
with the \textit{Adam} optimiser~\cite{Diederik2015} and a batch size of~16.
The autoencoder was trained for 25 epochs and returned an MSE loss of $2.9\times 10^{-4}$.
The model was trained for 50 epochs.
Unless explicitly stated otherwise,
all layer activation functions in the model are Leaky Rectified Linear Units~\cite{chollet2015keras,khalid2020}.

\subsection{Evaluation}
\label{subsec:evaluation}

We investigate training the model using $D=1$ (30 min)
and $D=4$ (2 hours) difference between initial and final probability maps.
In the former case, individual fires within the probability map may not significantly diverge,
and the behaviour may be difficult for the model to learn.
In the latter case, the dynamics are allowed more time to develop,
but this comes at the cost of more invocations of the inner model.
This recursive behaviour may lead to issues regarding vanishing/exploding gradients~\cite{hanin2018};
we found that the model weights failed to converge when using $D=8$.

We also evaluate two configurations (A and B) of the inner model (shown in Fig.~\ref{fig:arch_inner}).
In configuration A, layer $C$ acts as an identity operation,
simply passing through the latent probability map.
In configuration B, layer $C$ applies a Sobel edge filter with kernel size 3.
We also train the model under various crop window sizes \textit{c}.
The amount of padding \textit{p} is increased in proportion with the window size.
For example a cropped square of 256 pixels has a padding size of 32 pixels,
and a cropped square of 512 pixels has a padding size of 64 pixels.

We use two metrics to evaluate the performance: weighted MSE (WMSE) and Jaccard index.
The WMSE is the average of MSE weighted by the amount of area that is burned in each sample.
The area of land likely to be burned between initial and final states
is given by $A(y^k_t, y^k_0) = \sum y^k_t - y^k_0$ for a sample $k$.
We then define
$\text{WMSE}(y_t, y_p, y_0) = 1 / (\sum_k A(y^k_t, y^k_0)) \sum_k \text{MSE}(y^k_t, y^k_p) \times A(y^k_t, y^k_0)$,
where we sum over all $k$ samples. Lower values of WMSE indicate better performance.

The Jaccard index (aka Jaccard similarity coefficient)
is defined as the intersection between two sets divided by their union~\cite{duff2016}.
We use the Jaccard index to gauge how similar the probability maps directly generated by the proposed model
are to the probability maps indirectly generated via SPARK~\cite{Spark2015}
(ie., where an ensemble of simulations is flattened into a probability map),
with SPARK derived data treated as the ground-truth.
Higher values of the Jaccard index indicate higher similarity.
However, direct use of the Jaccard index on probability maps can be problematic,
as there is no implicitly defined boundary on which to evaluate the index.
As such, we use a threshold of 10\% to define the boundary of the target and prediction sets
(ie., probabilities less than 10\% are ignored).
This was chosen in order to evaluate the model's ability to capture the behaviour of reasonably likely events,
while not penalising the model for missing highly unexpected behaviour.
Furthermore, the 10\% threshold may meet the needs of fire risk management
(eg., decision planning and responsiveness),
where low likelihood events are less likely to be considered as actionable.

Results presented in Table~\ref{tab:evaluation} show that $D=4$ leads to better performance than $D=1$.
Furthermore, we observe a general reduction in the WMSE
as the size of the cropped region is increased from 256 to 512 pixels.
In terms of the Jaccard index,
there is a notable improvement with configuration B compared to configuration A,
which indicates that fire front detection via Sobel edge detection
is a useful feature for determining ensemble fire dynamics.

\begin{figure*}[!tb]
  \centering
  \begin{minipage}{0.3\textwidth}
    \centering
    \includegraphics[width=\textwidth]{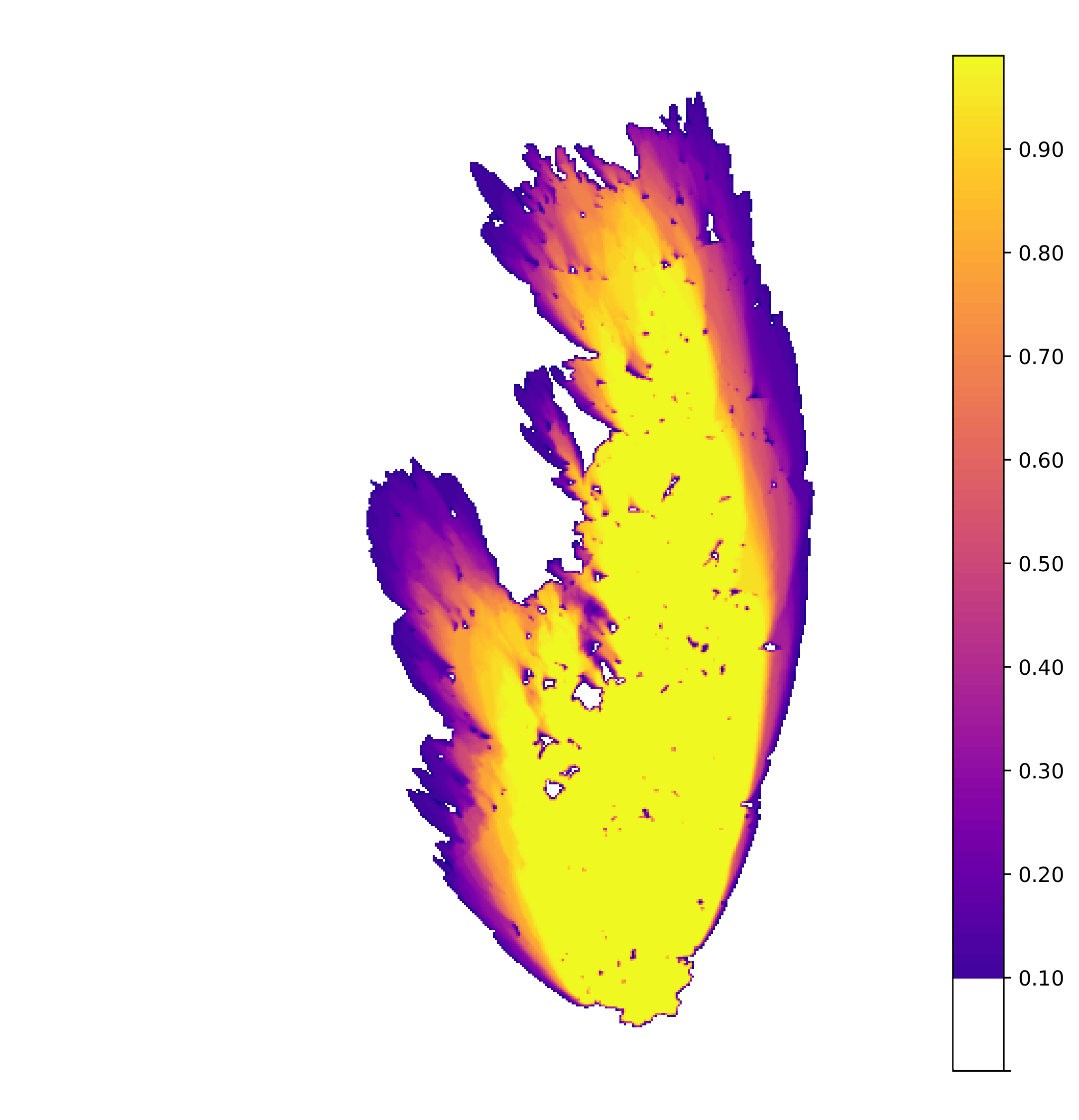}
    \footnotesize\textbf{(a)}\\
    \scriptsize Probability~map~from~ensemble~of~simulations
  \end{minipage}%
  \hfill
  \begin{minipage}{0.3\textwidth}
    \centering
    \includegraphics[width=\textwidth]{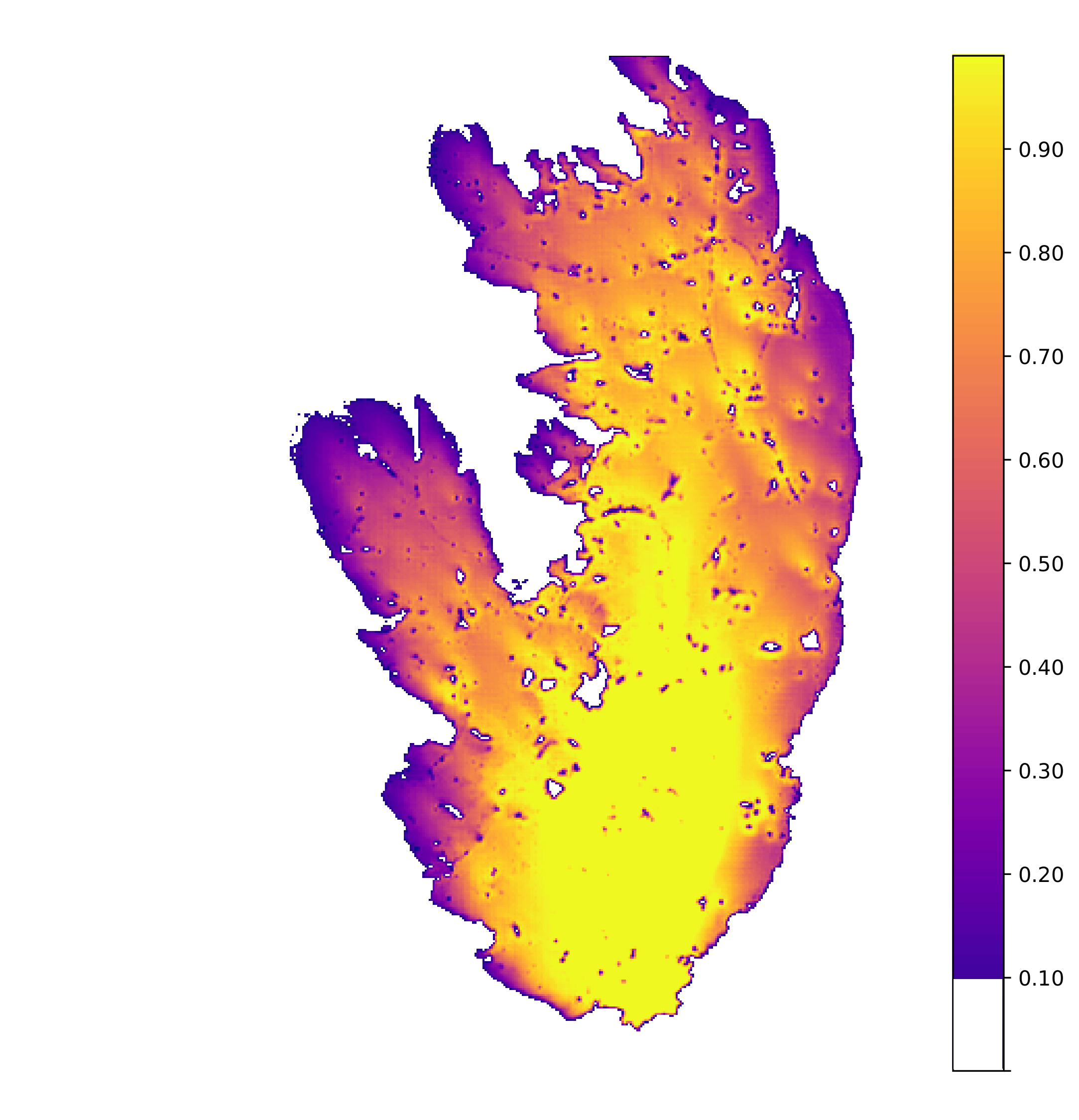}
    \footnotesize\textbf{(b)}\\
    \scriptsize Emulated probability map
  \end{minipage}%
  \hfill
  \begin{minipage}{0.3\textwidth}
    \centering
    \vspace{1ex}
    \includegraphics[width=\textwidth,height=0.95\textwidth]{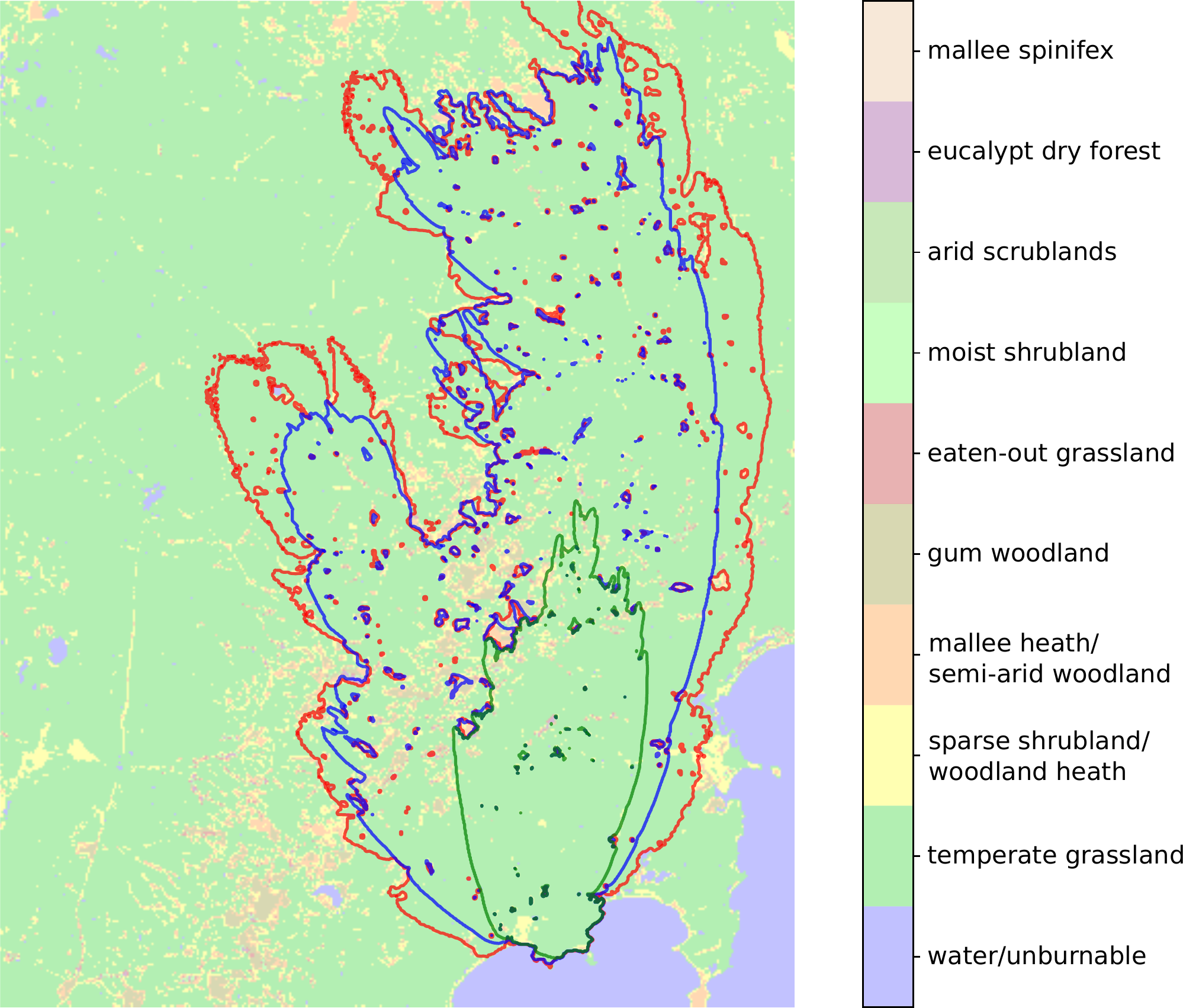}\\
    \vspace{1.5ex}
    \footnotesize\textbf{(c)}\\
    \scriptsize Initial perimeter and final contours
  \end{minipage}%
  \caption
    {
    \small
    \textbf{(a)}:
    A probability map that has been generated using an ensemble of 50 fires.
    The initial perimeter is marked as a green contour in panel~(c).
    From this starting location the fire burns for a duration of 14 hours.
    Each fire within the ensemble is generated using weather inputs that are subjected to random walk noise.
    The ensemble is then flattened to generate the probability map.
    Values less than 10\% are ignored.
    \textbf{(b)}:
    The corresponding probability map generated by the proposed model,
    with values less than 10\% ignored.
    The model uses summary weather variables (mean and standard deviation)
    to reproduce/emulate the ensemble shown in panel~(a).
    While the proposed model tends to overestimate high likelihood regions and underestimate low likelihood regions,
    the two maps qualitatively show a general agreement on shape.
    \textbf{(c)}:
    Contour lines for the probability maps at the 10\% threshold.
    Blue colour represents the contour in panel~(a),
    while red colour represents the contour in panel~(b).
    } 
  \label{fig:emulation_example_1}
  \vspace{1ex}
  \hrule
\end{figure*}

Fig.~\ref{fig:emulation_example_1} provides a graphical example of the performance
on a sample fire over a long duration (14 hours).
This example was produced using configuration B and a cropping size of 512 pixels during model training.
Likelihoods below the 10\% threshold have been truncated and are not shown.

\subsection{Comparison}
\label{sec:comparison}

As the proposed model is an evolution of the model presented in~\cite{bolt2022},
in this section we evaluate the quality of probability maps {directly} generated by the proposed model
(as per Section~\ref{subsec:evaluation})
against the quality of probability maps {indirectly} generated
from ensembles of emulated fires produced via the model in~\cite{bolt2022}.
In the latter case, 
the model is trained using noiseless weather variables on the same training dataset as the proposed model.

The quality is evaluated in terms of the Jaccard index
(with contours defined at the 10\% likelihood level),
where we measure how similar the probability maps are
to those indirectly generated via SPARK~\cite{Spark2015},
which are treated as ground-truth.
We also measure the time taken (in seconds) for each model to generate the probability maps,
running on the same hardware (a modern computer with a multicore CPU and high-end GPU)
and using the same software stack as per Section~\ref{subsec:metrics_and_loss}.
For the direct method we choose the configuration with the lowest training and evaluation loss,
which corresponds to setting shown in the last row of Table~\ref{tab:evaluation}.
For the indirect method, we vary the ensemble size for generating the probability maps;
three ensemble sizes are evaluated: 10, 20 and 50.

The results shown in Table~\ref{tab:evaluation2} indicate that
for an ensemble size of 50 (the same size as used for training the direct approach),
the indirect approach (ensembles of emulated fires produced via the model in~\cite{bolt2022})
produces somewhat higher quality maps.
However, this comes at the cost of requiring considerably longer execution time,
where the indirect approach is about 19 times slower than the direct approach.
For an ensemble size of 20,
the indirect approach produces probability maps which are roughly comparable in quality 
to the direct approach,
but is about 7 times slower.
For an ensemble size of 10,
the indirect approach produces considerably lower quality maps and is still noticeably slower.

Overall, for the indirect approach there is a correlation between execution time
and the quality of the resultant probability maps.
However, increasing the size of the ensemble from 20 to 50 only leads to a minor increase in quality,
at the cost of a large increase in the execution time.
The considerably lower execution time of the proposed direct approach
(while still producing reasonable quality probability maps)
allows for the evaluation of more fire progression scenarios within a given time budget.
This in turn can be useful for fire risk management,
where large ensembles are typically used to explore large variable spaces. 

\begin{table}[!t]
\begin{minipage}{1\textwidth}
\centering
\begin{minipage}{0.48\textwidth}
  \caption
    {
    \small
    Results in terms of WMSE and Jaccard index for various settings and configurations.
    Each setting comprises \textit{c} and \textit{p} components,
    where \textit{c} is size of the cropping window and \textit{p} is the amount of padded pixels.
    $D$ indicates the difference in time between initial and target probability maps used during training,
    measured in 30 minute intervals.
    Jaccard indices are calculated over the validation set for contours defined at the 10\% likelihood level.
    The indices indicate how similar the probability maps directly generated by the proposed model
    are to the probability maps indirectly generated via SPARK~\cite{Spark2015}.
    }
  \label{tab:evaluation}
  \vspace{1ex}
  \setlength{\tabcolsep}{5pt}
  \centering
  \small
  \begin{tabular}{lccccc}
    \toprule
    \makecell{\textbf{Setting~~~~~~~~~~}} &
    \makecell{\textbf{Training} \\ \textbf{WMSE}} &
    \makecell{\textbf{Validation} \\ \textbf{WMSE}}  &
    \makecell{\textbf{Jaccard}} \\
    \midrule
    \multicolumn{4}{l}{\textit{Config.~A (no edges), D=1}}\\
    256c, 32p & 0.0196 & 0.0211 & 0.528    \\ 
    384c, 48p & 0.0172 & 0.0187 & 0.532    \\
    512c, 64p & 0.0168 & 0.0173 & 0.561    \\
    \hline
    \multicolumn{4}{l}{\textit{Config.~A (no edges), D=4}}\\
    256c, 32p & 0.0145 & 0.0148 & 0.520    \\ 
    384c, 48p & 0.0138 & 0.0151 & 0.563    \\
    512c, 64p & 0.0136 & 0.0142 & 0.575    \\
    \hline
    \multicolumn{4}{l}{\textit{Config.~B (edges), D=4}}\\
    256c, 32p & 0.0129 & 0.0130 & 0.643    \\ 
    384c, 48p & 0.0101 & 0.0110 & 0.679    \\
    512c, 64p & 0.0087 & 0.0097 & 0.674    \\
    \bottomrule
  \end{tabular}
\end{minipage}
\hfill
\vline
\hfill
\begin{minipage}{0.48\textwidth}
  \caption
    {
    \small
    Comparison of neural-based methods for estimating fire probability maps
    in terms of the Jaccard index (with contours defined at the 10\% likelihood level)
    on the validation set,
    using SPARK~\cite{Spark2015} derived maps as ground-truth.
    Two methods are evaluated:
    ensembles of emulated fires produced via the model in~\cite{bolt2022},
    and the proposed model which directly generates probability maps.
    The proposed model uses the same setting as shown in the last row of Table~\ref{tab:evaluation}.
    Execution time is reported in two forms:
    number of seconds, 
    and relative to the execution time of the direct approach.
    }
  \label{tab:evaluation2}
  \vspace{1ex}
  \setlength{\tabcolsep}{5pt}
  \centering
  \small
  \begin{tabular}{lcccc}
    \toprule
    \makecell{\textbf{Method}} &
    \makecell{\textbf{Ensemble}\\\textbf{Size}} &
    \makecell{\textbf{Jaccard}} & 
    \makecell{\textbf{time}\\ \textbf{(sec.)}} &
    \makecell{\textbf{time}\\ \textbf{(rel.)}} \\
    \midrule
    \multirow{3}{*}{\makecell{ensembles of\\emulated fires}} & 10 & 0.468 & ~~8.6 & ~~3.7 \\
    ~                                       & 20 & 0.707 &  17.6 & ~~7.6 \\
    ~                                       & 50 & 0.733 &  44.0 &  19.0 \\
    \midrule
    \makecell{direct\\probability map} & \makecell{~\\--\\~}  & 0.674 & 2.31 & 1.0 \\
    \bottomrule
  \end{tabular}
  \vspace{16.5ex}
\end{minipage}
\end{minipage}
\vspace{-1ex}
\end{table}

\newpage
\section{Concluding Remarks}
\label{sec:conclusion}

In this paper we have demonstrated that deep neural networks can be constructed
to directly generate probability maps for estimating wildfire spread.
This direct approach can be used instead of indirectly obtaining probability maps via ensembles of simulations,
which are typically computationally expensive.
In comparison to ensembles of simulated fires obtained via the traditional SPARK platform~\cite{Spark2015},
the proposed approach achieves reasonable quality probability maps
when weather forecasts are exposed to random walk noise.
On an evaluation set of 35 fires, the overall Jaccard index (similarity score) is $67.4\%$.
Qualitatively, the proposed approach reproduces the correct general response to fuel class and wind direction,
and the shape of the probability maps is comparable to those produced by traditional simulation.

The computational load for the proposed approach is concentrated in the model training process,
rather than during deployment.
When compared to a related neural model (emulator) which was employed
to generate probability maps via ensembles of emulated fires,
the proposed model produces competitive Jaccard similarity scores
while being considerably less computationally demanding.

We observe that the model does not generate a perfectly smooth transition of probabilities from high to low,
as observed in probability maps from ensembles of simulated fires.
This will likely limit the capability for the model to predict accurately over long durations.
Further refinement of the architecture, model restraints, summary variables and incorporation of variable distributions
may yield better results. 

A further area of research would be the incorporation of better noise models for weather variables.
In this work we have used a straightforward noise model, aimed for demonstrating proof-of-concept.
This noise model was not developed to be representative of the complexities in weather forecasting.
A more detailed discussion of more comprehensive weather models
that incorporate uncertainty can be found in~\cite{murphy1985, palmer2000, slingo2011}.
In our noise model, any two fires within an ensemble are likely to smoothly diverge from each other over time,
except in cases where natural barriers cause bifurcating behaviour.
In more realistic settings, bifurcation may occur due to rapid changes in wind direction
that are predicted to occur at various times within an ensemble of simulations.
It may be possible to capture the behaviour of more complex noise models
by using binning methods to generate summary variables,
rather than simply using mean and standard deviation.

More sophisticated fire spread models
(for example ones that incorporate ember attacks~\cite{Sharifian2016,wang2006})
may require very large ensembles in order to reliably gauge the behaviour of fires.
Generating likelihood estimates through very large ensembles is likely to be computationally taxing.
Direct generation of probability maps, as proposed in this work, may be very beneficial in these contexts. 
Finally, we note that the proposed approach is not specific to wildfire prediction,
and other similar spatio-temporal problems may benefit from direct likelihood estimation,
such as diffusion of pollution, pest and disease spread, and flood impact models. 

~

\small
\noindent
\textbf{Acknowledgements.}
We would like to thank our colleague Peter Wang for providing results from the SPARK simulation platform.

\small
\bibliographystyle{ieee_mod}
\bibliography{references}

\end{document}